\documentclass[conference]{IEEEtran}
\IEEEoverridecommandlockouts
\usepackage{cite}
\usepackage{amsmath,amssymb,amsfonts}
\usepackage{algorithmic}
\usepackage{graphicx}
\usepackage{textcomp}
\usepackage{url}
\usepackage{xcolor}
\def\BibTeX{{\rm B\kern-.05em{\sc i\kern-.025em b}\kern-.08em
    T\kern-.1667em\lower.7ex\hbox{E}\kern-.125emX}}

\usepackage{multirow}
\usepackage{subcaption}
\usepackage{censor}
\usepackage{float}
\usepackage{balance}

\usepackage[pscoord]{eso-pic}
\newcommand{\placetextbox}[3]{
\setbox0=\hbox{#3}
\AddToShipoutPictureFG{ \put(\LenToUnit{#1\paperwidth},\LenToUnit{#2\paperheight}){\vtop{{\null}\makebox[0pt][c]{#3}}}
}
}
\placetextbox{.5}{0.060}{\footnotesize { \copyright Authors. Personal use of this material is permitted. Permission must be obtained for all other uses, in any current or future media } }
\placetextbox{.5}{0.050}{\footnotesize {including reprinting/republishing this material for advertising or promotional purposes, creating new  collective works, }}
\placetextbox{.5}{0.040}{\footnotesize {for resale or redistribution to servers or lists, or reuse of any copyrighted component of this work in other works.}}

\begin{document}

% \title{Performance Characterization of Large Language Models on Recent Resource-Constrained Hardware}
\title{\huge {\bf Large Language Models on Small Resource-Constrained Systems: Performance Analysis and Trade-offs}}

 \author{\IEEEauthorblockN{
\large Liam Seymour\\
 \normalsize Electrical Engineering, Computer Science\\
 \normalsize Western Kentucky University, KY, USA\\
 %\normalsize Bowling Green, Kentucky, USA\\
 \normalsize william.seymour656@topper.wku.edu}
 \vspace{-4mm}
 \and
 \IEEEauthorblockN{
 \large Basar Kutukcu\\
 \normalsize Electrical and Computer Engineering\\
 \normalsize University of California San Diego, CA, USA\\
 %\normalsize {\color{blue} (TODO: Add City)}\\
 \normalsize bktkc@ucsd.edu}
 \vspace{-4mm}
 \and
 \IEEEauthorblockN{
 \large Sabur Baidya\\
 \normalsize Computer Science and Engineering\\
 \normalsize University of Louisville, KY, USA\\
 %\normalsize Louisville, Kentucky, USA\\
 \normalsize sabur.baidya@louisville.edu}
 \vspace{-4mm}
 %{\color{blue} (TODO: is this the right order?)}
 }

% \author{\IEEEauthorblockN{
% \large Researcher A\\
% \normalsize Electrical Engineering, Computer Science\\
% \normalsize University A\\
% \normalsize Location A\\
% \normalsize researcher.a@email.edu}
% \vspace{-4mm}
% \and
% \IEEEauthorblockN{
% \large Researcher B\\
% \normalsize Electrical and Computer Engineering\\
% \normalsize University B\\
% \normalsize Location B\\
% \normalsize researcher.b@email.edu}
% \vspace{-4mm}
% \and
% \IEEEauthorblockN{
% \large Researcher C\\
% \normalsize Computer Science and Engineering\\
% \normalsize University C\\
% \normalsize Location C\\
% \normalsize researcher.c@email.edu}
% \vspace{-4mm}
% }

\maketitle

\begin{abstract}

Generative AI like the Large Language Models (LLMs) has become more available for the general consumer in recent years. Publicly available services, e.g., ChatGPT, perform token generation on networked cloud server hardware, effectively removing the hardware entry cost for end users. However, the reliance on network access for these services, privacy and security risks involved, and sometimes the needs of the application make it necessary to run LLMs locally on edge devices. A significant amount of research has been done on optimization of LLMs and other transformer-based models on non-networked, resource-constrained devices, but they typically target older hardware.
Our research intends to provide a `baseline' characterization of more recent commercially available embedded hardware for LLMs, and to provide a simple utility to facilitate batch testing LLMs on recent Jetson hardware. We focus on the latest line of NVIDIA Jetson devices (Jetson Orin), and a set of publicly available LLMs (Pythia) ranging between 70 million and 1.4 billion parameters. Through detailed experimental evaluation with varying software and hardware parameters, we showcase trade-off spaces and optimization choices.
Additionally, we design our testing structure to facilitate further research that involves performing batch LLM testing on Jetson hardware.
\end{abstract}

% TODO: replace with ACM keywords after ACM format shift
\begin{IEEEkeywords}
Embedded Systems, Large Language Models, Evaluation, Machine Learning, Resource Constraints, Characterization, Performance Benchmark
\end{IEEEkeywords}

\section{Introduction}

Generative AI, e.g., the large language model (LLM)~\cite{feuerriegel2024generative, zhao2023survey}, has become a prevalent service for general consumer use and many other modern applications~\cite{fui2023generative,wang2023pre}. Widely known standalone cloud-based services like ChatGPT~\cite{roumeliotis2023chatgpt} can provide chat/text generation for free to consumers and hence, are considerably appealing due to the complexity of running LLMs.
% why do we care about running LLMs locally?
Despite this, local execution of LLMs is an important field of study as many modern applications, e.g., healthcare and robotic applications, started using LLMs for processing sensitive information which need to be executed locally instead of using cloud-based generative AI services. Additionally, 
there are several limitations and undesirable side effects in the network used for the cloud-based services, e.g., communication failure due to bad connectivity or mobility of the device used for the applications. There is also security risk in processing any sensitive information through cloud-based LLM services as well. Hence executing the LLMs for these applications on the edge devices have become increasingly necessary and relevant.

However, executing the LLMs on-device has some immediate drawbacks -- namely the hardware requirements for larger models. Much of the speed of LLM text generation comes from the fact that these servers have plenty of acceleration hardware available. Consumer devices with high-performing compatible GPU hardware may be able to achieve similar performance; however, performance can still be limited by the available video memory on these devices. Models that are too large to be loaded in entirety suffer from the ``memory wall'' problem \cite{xu2023llmcad} and must have weights loaded/unloaded mid-generation to function.

% why jetsons
Embedded systems are in a tier below consumer hardware, but recent advancement in embedded hardware provides some options for running LLMs on these resource-constrained devices.
NVIDIA provides a family of devices (Jetson)~\cite{archet2023embedded} with their GPU hardware built into a system-on-a-chip (SoC) and varied amounts of memory. 
%These Jetson devices are of particular interest for embedded LLMs because they contain 
Several recent works on LLMs/Transformer models utilize Jetson devices for their on-board GPU hardware. Some studies focus on the previous line of devices (TX2, Xavier) \cite{schubert2023}\cite{tabani2021}, while others introduce a single device configuration from the newest line (known as Orin) \cite{xu2023llmcad} in a variety of tests.

% in this paper we
The goal of this paper is to explore the process of performing a full range of tests on LLMs on these dynamically configurable Jetson Orin devices, as well as providing our created testing setup and findings to assist with future research with these devices. In particular, we target a variety of hardware configuration options (i.e. Orin device config, NV power model) and software options for LLMs (i.e. LLM parameter size, quantization). Finally, we showcase a trade-off space in hardware and software configurations for different optimization objectives and performance constraints.

\section{Related Work}
%As mentioned above, 
NVIDIA Jetson boards are popular choices in several research works for on-device generation and characterization \cite{jetson_nano_benchmark1}\cite{jetson_dl_tensorrt}\cite{tx2_dnn_imagenet}.
Other studies have been conducted on the Orin line of devices, especially for other non-transformer-based deep-learning models \cite{archet2023embedded}. Some other works, e.g., \cite{EvoSh} used
older generation of Jetson devices 
%are not irrelevant for research -- 
to find the optimized configurations of deep-learning models and hardware parameters for execution of deep learning applications.
%are a subject of recent research utilizing the previous generation of Jetson devices .

Jetson devices are not the only devices used by the researchers; many other ARM-based embedded boards are also common targets for testing models \cite{dnns_arm_devs}.
Smartphones are another common device for on-device models, as they offer reasonable GPU resources for heterogeneous computing \cite{rstensorflow_android_gpu}.
Additional work also exists for hardware implementations (FPGA) of LLMs, which help subvert conventional issues with GPU-enabled computation (i.e. memory overhead, heavy computation, low cost efficiency) \cite{FPGA_LLM1}.

Fully on-device generation/computation is but one of multiple use cases for deep learning models on embedded devices \cite{edge_computing_review}.
Edge server offloading is another viable method to perform model computation, and reduction of the latency tradeoff of utilizing networked server resources has been a subject of research \cite{liu2019edgeserver}.
This also enables edge devices to intelligently choose deep-learning models for offloaded computation based on specific constraints \cite{ran2018deepdecision}.

Making LLMs more available for on-device processing by improving their efficiency has been an active research area as well~\cite{efficient_LLM_Survey}.
Partially explored in this research, quantization is a common research topic for successful LLM deployment on resource-constrained devices, and can show significant performance advantages \cite{lin2024quantization}.
Pruning is another common research area for LLMs, focusing on various techniques of intelligent reduction of model parameters \cite{xia2024pruning}. In this paper, we explored different hardware configurations of the embedded computing unit and also, software configurations, e.g., model hyperparameters to measure the benchmark characterization and analyze the trade-off space for optimized co-design choices.

\vspace{4mm}
\section{Approach}

To accumulate data for our characterization, we utilized the Jetson Orin developer kit and its ability to match the hardware features and performance of other Jetson Orin devices. By design, the Orin devices all ``{share one SoC architecture, enabling the developer kit to emulate performance and power for any of the [devices]}''\cite{orinuserguide}. Flashing different configurations to the developer kit enables/disables different hardware (CPU and GPU cores, available memory, etc.) to perform this emulation. This allowed us to fully examine several devices in the Orin line (as well as the performance of the development kit in the default configuration). In particular, Table \ref{table:dev_info} shows a list of our targeted devices.

Our tests included a set of five Pythia LLM models with a varied number of model parameters. These models are publicly available and constructed on the same dataset, with the intent to be used in research into the effect of model scalability\cite{biderman2023}. Models were made available by the HuggingFace suite \cite{hftransformers} and have been uploaded with 16-bit parameter precision. 

\subsection{Experimentation}

We obtained and recorded the following metrics in our characterization tests:
\begin{itemize}
    \item \textit{Latency} -- How long does the LLM take to load and complete token generation?
    \item \textit{Power} -- How much power is used by the LLM during loading and generation?
    \item \textit{Memory} -- How much memory (RAM, GPU) is used by the LLM during loading and generation?
    \item \textit{Accuracy} -- How accurate is the LLM model itself?
\end{itemize}

Using these metrics, we calculated estimates for the following derived metrics:
\begin{itemize}
    \item \textit{Energy} -- What is the estimated additional energy usage by the LLM during generation?
    \item \textit{Time per Token} -- How long does it take (on average) to generate a single token?
\end{itemize}

We varied the following testing parameters and performed an exhaustive sweep on the different configurations:
\begin{enumerate}
    \item \textit{Device Configuration} -- As mentioned, the Orin developer kit was flashed between six (6) different possible configurations. These configurations are shown in Table \ref{table:dev_info} along with some GPU information. 
    \item \textit{LLM} -- Five (5) of the Pythia\cite{biderman2023} models were used, ranging from 70 million parameters (pythia-70m-deduped) to 1.4 billion parameters (pythia-1.4b-deduped).
    \item \textit{NV Power Model} -- For each device configuration, tests were completed for each of the default power models provided. While it is possible to make custom power models for specific cases, these come enabled with default JetPack versions for the devices and are more likely to be selected by end users.
    \item \textit{Quantization} -- For each LLM, tests were conducted both with 4-bit quantization of parameters and without any quantization. The Pythia models are provided 
\end{enumerate}

Additionally, each test configuration was performed $5$ times sequentially before moving to the next. This was to allow us to compare the behavior of the initial run to subsequent loading and generation.

% device table
\begin{table}[]
    \centering
    \begin{tabular}{|c|c|c|}
         \hline
         \textbf{Device} & \textbf{CUDA Cores} & \textbf{Unified Memory \cite{unified_memory}} \\
         \hline
         AGX Orin Devkit & 2048 & 32 GB \\
         AGX Orin 32GB & 1792 & 32 GB \\
         Orin NX 16GB & 1024 & 16 GB \\
         Orin NX 8GB & 1024 & 8 GB \\
         Orin Nano 8GB & 1024 & 8 GB \\
         Orin Nano 4GB & 512 & 4 GB \\
         \hline
    \end{tabular}
    \vspace{2ex}
    \caption{List of targeted Jetson Orin devices and some additional information\cite{orinspecs}. Note that at the time of writing, 64 GB models of the Devkit are available; the device in this study is a slightly older model with 32 GB of memory.}
    \label{table:dev_info}
\end{table}

\subsection{Implementation}

Our testing suite was developed in Python to facilitate use of the HuggingFace suite, and utilized PyTorch as the underlying implementation for the models used. PyTorch was chosen due to compatibility with the majority of models on the HuggingFace hub, which would assist further testing using our suite on different LLMs.

% isolation (multiprocessing, multi-user.target, ssh, no monitor connected)
To isolate our testing as much as possible and to reduce extraneous processes from using resrou
All individual tests are run in order on separate processes, spawned from the main logging process.

% periods
Our main testing script divides measurement into three (3) periods with the given state flags:
\begin{enumerate}
    \item IDLE. A 15-second interval where nothing happens, but the subprocess is running and the necessary Python libraries are loaded. This allows us to measure a ``baseline'' power estimate. 
    \item MODEL\textunderscore LOAD. The period in which the LLM is loaded into GPU memory. If specified, the model can be loaded with or without 4-bit quantization.
    \item GENERATE. The period in which the LLM is tasked with generating a given number of tokens. The period ends when the LLM is finished generating these tokens.
\end{enumerate}

\begin{figure*}[h]
    \centering
    {\begin{subfigure}[b]{0.49\textwidth}
        \centering
        \includegraphics[width=\textwidth]{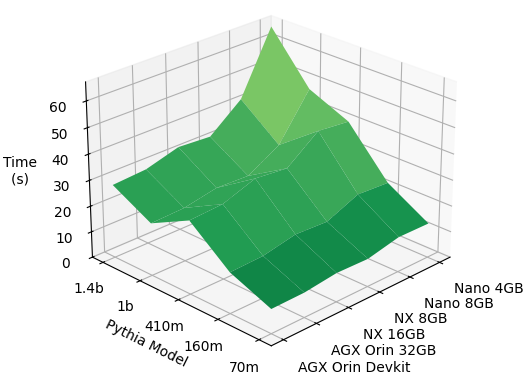}
        \vspace{-1em}
        \caption{Median total token generation latency (in seconds), for the maximum NV power model, with 4-bit quantization.}
        \label{fig:3d_gen_q}
    \end{subfigure}
    \hfill
    \begin{subfigure}[b]{0.49\textwidth}
        \centering
        \includegraphics[width=\textwidth]{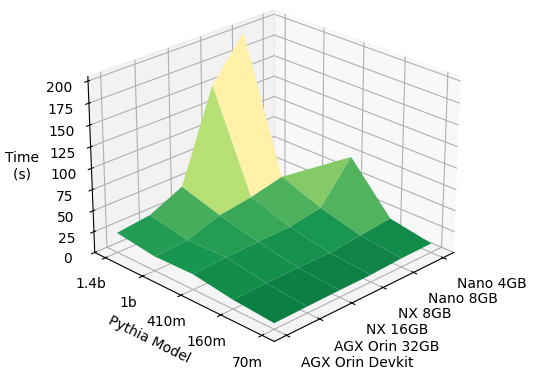}
        \vspace{-1em}
        \caption{Median total token generation latency (in seconds), for the maximum NV power model, without quantization.}
        \label{fig:3d_gen_nq}
    \end{subfigure}}
    {\begin{subfigure}[b]{0.49\textwidth}
        \centering
        \includegraphics[width=\textwidth]{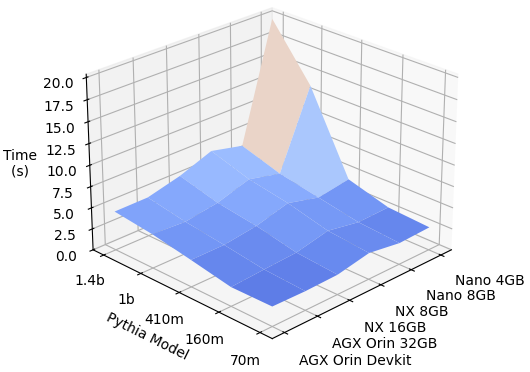}
        \vspace{-1em}
        \caption{Median model loading latency (in seconds), for the maximum NV power model, with 4-bit quantization.}
        \label{fig:3d_load_q}
    \end{subfigure}
    \hfill
    \begin{subfigure}[b]{0.49\textwidth}
        \centering
        \includegraphics[width=\textwidth]{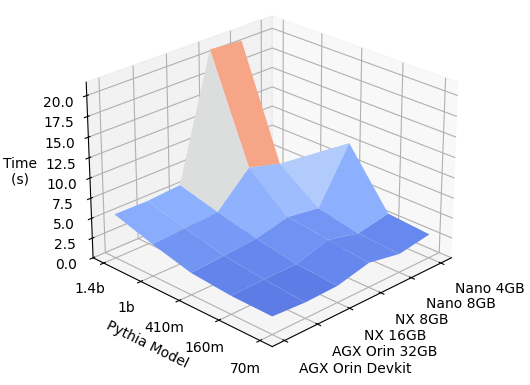}
        \vspace{-1em}
        \caption{Median model loading latency (in seconds), for the maximum NV power model, without quantization.}
        \label{fig:3d_load_nq}
    \end{subfigure}}
    \caption{Latency results across all device configurations, for both model loading and token generation.}
    \label{fig:3d_gen}
    \vspace{-4mm}
\end{figure*}

% metrics
The individual metric implementations – as well as the derived metrics – are described in the following subsections:

{\bf \textit{Latency:}}
Timestamps are made in the log for the beginning and end of each of the three periods mentioned above. These timestamps are named with the state flag name, followed by either ``\_START'' or ``\_END'' for pair identification in the analysis scripts. Our logging system uses Python's built-in time.perf\_counter() with the timestamps, such that the amount of time elapsed during a period is equal to the difference between the time values of the timestamp pair.

{\bf \textit{Power:}}
The most reliable method we found for measuring estimated power usage on Jetson devices is through the Python library ``jetson-stats''\cite{jetsonstats}. This library allows us to poll many Jetson-specific metrics at regular intervals through a systemd process named `jtop'. To measure power usage, we use the built-in total power usage metric, which reads voltage and current directly from the internal rails via the PMIC (power management IC) on the developer kit. After reconfiguring the jtop server from default settings, we were able to reach (on average) roughly a ¼-second measurement interval (with the minimum interval being 0.1 seconds), and these measurements are taken regardless of the current test state.  

{\bf \textit{Memory:}}
The jtop method is also capable of measuring memory statistics, though the memory is always associated with a specific running process. The jtop server automatically keeps track of which processes have some stake in the GPU memory, and provides access to connected Python scripts\cite{jetsonstats}. By default, our logging implementation stores memory usage for all processes that use PyTorch in some way. Since we were not testing anything in parallel and reducing as much GPU usage as possible outside of our tests, logs typically only reflect a single process using GPU memory. Both GPU memory usage and RAM usage are logged for the process(es) found.

{\bf \textit{Accuracy:}}
This metric was gathered separately from the others, using the LM evaluation harness provided by EleutherAI\cite{eval-harness} instead of our primary testing suite. This software is meant to perform accuracy measurements on various LLMs on different platforms. These accuracy measurements are independent from the hardware, being more dependent on the structure of the model itself. Because of this, we were able to obtain a single set of measurements on each LLM (for the given quantization levels) on remote hardware. While this is not indicative of some specific feature of the Jetson hardware, it does provide a frame of reference for the models we chose to use in this characterization.

{\bf \textit{Energy:}}
As we obtain many samples of the average power of the Jetson via jtop, we are able to roughly estimate the energy usage of the device during model loading and token generation. By using the median power in the IDLE period as a ``baseline'' power measurement for when the device is not using any models, we can subtract this value from each sample in either of the other periods to get time-series data for the \textit{additional power} required to load and/or generate. By integrating this time-series data, we are able to obtain the (estimated) \textit{additional energy} required to load and/or generate, according to the physical relation between electrical energy and power. We implemented this in our analysis scripts with a simple trapezoidal integration.

{\bf \textit{Time per Token:}}
We were able to obtain the average generation time per token by dividing the final generation time for each test by the number of tokens generated. While it is only an average, this provides insight into the scalability of performing generation on each device, i.e. how well a particular device may work on a much larger (or a much smaller) desired output. By testing with a large number of desired tokens, we better approximate this metric.
At the time of writing, our log data only includes tests with a set number of tokens always being generated (512). However, our analysis scripts are capable of working with varied numbers of tokens, and the number at the end of generation is saved with the log.

\subsection{Deployment}

\begin{table*}[]
    \centering
    \begin{tabular}{|c|c|c|c|c|c|c|c|c|c|c|c|}
        \cline{1-12}
        \multicolumn{2}{|c}{\multirow{2}{*}{\textbf{Median Model Loading Latency (s)}}} & \multicolumn{10}{|c|}{Pythia LLM} \\
        \cline{3-12}
        %\multicolumn{2}{c}{} & \multicolumn{2}{|c|}{70m} & \multicolumn{2}{|c|}{160m} & \multicolumn{2}{|c|}{410m} & \multicolumn{2}{|c|}{1b} & \multicolumn{2}{|c|}{1.4b} \\
        \multicolumn{2}{|c}{} & \multicolumn{5}{|c|}{4-bit Quantization} & \multicolumn{5}{|c|}{No Quantization} \\
        \hline
        Device & Power Model 
        & 70m & 160m & 410m & 1b & 1.4b
        & 70m & 160m & 410m & 1b & 1.4b \\
        \hline
        \multirow{4}{*}{AGX Orin Devkit}
        & MAXN & 2.280 & 2.229 & 3.142 & 4.002 & 4.568 & 2.182 & 2.272 & 2.722 & 4.000 & 5.502 \\
        & 50W & 2.503 & 2.880 & 3.922 & 4.679 & 5.231 & 2.362 & 2.554 & 3.414 & 4.616 & 6.038 \\
        & 30W & 2.376 & 2.672 & 4.049 & 4.438 & 5.828 & 2.100 & 2.412 & 3.014 & 4.149 & 6.103 \\
        & 15W & 3.809 & 4.336 & 5.728 & 7.182 & 8.341 & 3.829 & 4.419 & 5.263 & 7.394 & 9.648 \\
        \hline
        \multirow{4}{*}{AGX Orin 32GB} 
        & MAXN & 2.208 & 2.480 & 3.167 & 3.776 & 4.606 & 1.943 & 2.390 & 3.018 & 4.196 & 5.450 \\
        & 40W & 2.552 & 3.044 & 3.850 & 4.262 & 5.462 & 2.195 & 2.672 & 3.383 & 4.548 & 6.023 \\
        & 30W & 2.400 & 2.721 & 3.791 & 4.464 & 5.590 & 2.132 & 2.583 & 3.250 & 4.340 & 5.709 \\
        & 15W & 3.704 & 4.017 & 5.363 & 6.400 & 7.947 & 3.003 & 3.985 & 5.010 & 6.883 & 9.191 \\
        \hline
        \multirow{4}{*}{Orin NX 16GB} 
        & MAXN & 2.314 & 2.701 & 3.699 & 4.376 & 5.585 & 2.073 & 2.421 & 3.355 & 4.482 & 5.767 \\
        & 25W & 2.600 & 2.916 & 3.880 & 4.833 & 6.046 & 2.322 & 2.696 & 3.716 & 4.796 & 5.981 \\
        & 15W & 3.271 & 3.764 & 5.166 & 5.953 & 6.992 & 2.860 & 3.600 & 5.129 & 6.055 & 6.967 \\
        & 10W & 3.519 & 4.126 & 5.518 & 6.468 & 7.436 & 3.013 & 3.751 & 5.329 & 6.240 & 7.319 \\
        \hline
        \multirow{4}{*}{Orin NX 8GB} 
        & MAXN & 3.083 & 3.551 & 4.748 & 5.582 & 7.034 & 3.049 & 3.625 & 4.290 & 8.397 & 21.178 \\
        & 20W & 3.482 & 3.677 & 5.209 & 5.621 & 7.524 & 2.917 & 3.506 & 4.886 & 8.658 & 21.617 \\
        & 15W & 3.294 & 3.785 & 5.174 & 6.066 & 7.455 & 3.015 & 3.515 & 4.853 & 10.348 & 21.168 \\
        & 10W & 4.399 & 4.414 & 5.365 & 6.295 & 8.268 & 3.250 & 3.769 & 5.107 & 10.514 & 24.082 \\
        \hline
        \multirow{2}{*}{Orin Nano 8GB} 
        & 15W & 2.525 & 2.889 & 3.786 & 4.679 & 6.135 & 2.318 & 2.648 & 3.673 & 7.264 & 20.862 \\
        & 7W & 6.211 & 6.640 & 7.741 & 8.581 & 10.880 & 5.367 & 5.645 & 7.134 & 10.812 & 25.962 \\
        \hline
        \multirow{3}{*}{Orin Nano 4GB} 
        & 10W & 2.693 & 3.005 & 4.385 & 13.761 & 19.990 & 2.970 & 3.405 & 10.284 & - & - \\
        & 7W-AI & 5.364 & 5.829 & 8.029 & 17.088 & 24.308 & 4.762 & 5.260 & 13.513 & - & - \\
        & 7W-CPU & 4.580 & 5.376 & 6.887 & 16.520 & 25.060 & 5.251 & 6.656 & 12.849 & - & - \\
        \hline
    \end{tabular}
    \vspace{2ex}
    \caption{Median model loading latency (in seconds) across all LLMs, devices, and NV power models.}
    \label{table:lat_load}
\end{table*}

\begin{table*}[]
    \centering
    \begin{tabular}{|c|c|c|c|c|c|c|c|c|c|c|c|}
        \cline{1-12}
        \multicolumn{2}{|c}{\textbf{Median Total}} & \multicolumn{10}{|c|}{Pythia LLM} \\
        \cline{3-12}
        %\multicolumn{2}{c}{} & \multicolumn{2}{|c|}{70m} & \multicolumn{2}{|c|}{160m} & \multicolumn{2}{|c|}{410m} & \multicolumn{2}{|c|}{1b} & \multicolumn{2}{|c|}{1.4b} \\
        \multicolumn{2}{|c}{\textbf{Token Generation Latency (s)}} & \multicolumn{5}{|c|}{4-bit Quantization} & \multicolumn{5}{|c|}{No Quantization} \\
        \hline
        Device & Power Model 
        & 70m & 160m & 410m & 1b & 1.4b
        & 70m & 160m & 410m & 1b & 1.4b \\
        \hline
        \multirow{4}{*}{AGX Orin Devkit}
        & MAXN & 9.279 & 15.639 & 28.085 & 20.171 & 28.229 & 7.033 & 10.840 & 18.021 & 17.157 & 23.596 \\
        & 50W & 12.831 & 21.878 & 39.781 & 28.453 & 40.423 & 10.037 & 15.980 & 27.348 & 21.557 & 29.324 \\
        & 30W & 11.877 & 20.430 & 36.870 & 26.231 & 37.386 & 8.164 & 12.890 & 22.313 & 29.961 & 44.196 \\
        & 15W & 18.068 & 31.319 & 56.010 & 45.161 & 63.684 & 14.155 & 22.192 & 39.728 & 50.363 & 75.584 \\
        \hline
        \multirow{4}{*}{AGX Orin 32GB} 
        & MAXN & 9.475 & 16.104 & 28.874 & 20.676 & 29.102 & 7.744 & 12.108 & 20.522 & 19.378 & 28.210 \\
        & 40W & 12.844 & 21.975 & 39.806 & 28.364 & 39.989 & 9.230 & 14.376 & 24.613 & 21.929 & 31.808 \\
        & 30W & 11.123 & 19.090 & 34.242 & 25.156 & 35.675 & 8.446 & 13.464 & 23.276 & 30.009 & 44.199 \\
        & 15W & 16.331 & 28.151 & 49.299 & 44.810 & 63.337 & 11.824 & 19.140 & 35.622 & 50.015 & 75.109 \\
        \hline
        \multirow{4}{*}{Orin NX 16GB} 
        & MAXN & 11.045 & 18.681 & 33.327 & 23.150 & 32.704 & 8.316 & 12.456 & 21.537 & 32.037 & 46.230 \\
        & 25W & 13.044 & 22.444 & 40.343 & 36.070 & 51.132 & 10.481 & 17.609 & 38.219 & 66.690 & 98.056 \\
        & 15W & 15.009 & 23.343 & 41.613 & 43.268 & 62.216 & 11.141 & 17.789 & 66.613 & 70.130 & 106.109 \\
        & 10W & 16.881 & 27.116 & 47.865 & 43.771 & 62.810 & 11.900 & 19.155 & 67.002 & 74.283 & 112.441 \\
        \hline
        \multirow{4}{*}{Orin NX 8GB} 
        & MAXN & 10.719 & 18.067 & 32.059 & 22.679 & 31.853 & 8.463 & 12.584 & 22.514 & 37.568 & 152.470 \\
        & 20W & 13.965 & 24.185 & 43.659 & 36.191 & 51.242 & 10.708 & 17.843 & 38.253 & 66.857 & 198.364 \\
        & 15W & 14.772 & 22.725 & 40.413 & 43.250 & 62.252 & 10.149 & 16.730 & 66.526 & 69.851 & 208.090 \\
        & 10W & 17.122 & 27.726 & 48.494 & 43.866 & 62.821 & 12.343 & 19.708 & 67.112 & 73.803 & 247.655 \\
        \hline
        \multirow{2}{*}{Orin Nano 8GB} 
        & 15W & 13.470 & 23.262 & 41.447 & 29.787 & 41.838 & 10.406 & 16.261 & 28.700 & 46.294 & 200.976 \\
        & 7W & 21.284 & 35.029 & 57.869 & 63.192 & 90.568 & 15.820 & 25.970 & 98.427 & 104.519 & 306.432 \\
        \hline
        \multirow{3}{*}{Orin Nano 4GB} 
        & 10W & 13.273 & 22.523 & 40.199 & 46.996 & 65.859 & 10.911 & 19.835 & 74.400 & - & - \\
        & 7W-AI & 25.500 & 39.932 & 71.302 & 66.015 & 93.948 & 17.082 & 27.900 & 103.493 & - & - \\
        & 7W-CPU & 21.998 & 35.049 & 63.432 & 82.631 & 119.013 & 16.441 & 31.525 & 130.609 & - & - \\
        \hline
    \end{tabular}
    \vspace{2ex}
    \caption{Median token generation latency (in seconds) across all LLMs, devices, and NV power models.}
    \label{table:lat_gen}
    \vspace{-4mm}
\end{table*}

\begin{figure}[h]
    \centering
    \vspace{-1mm}
    \includegraphics[trim={0 0 0 1.2cm},clip,width=0.99\linewidth]{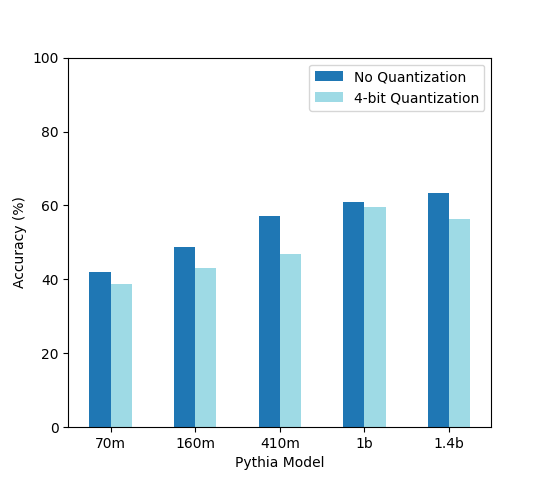}
    \vspace{-6mm}
    \caption{Accuracy of each LLM, tested using the LM Evaluation Harness\cite{eval-harness}.}
    \label{fig:accuracy}
    \vspace{-6mm}
\end{figure}

\begin{figure}[]
    \centering
    \includegraphics[width=0.95\linewidth, height=6.5cm]{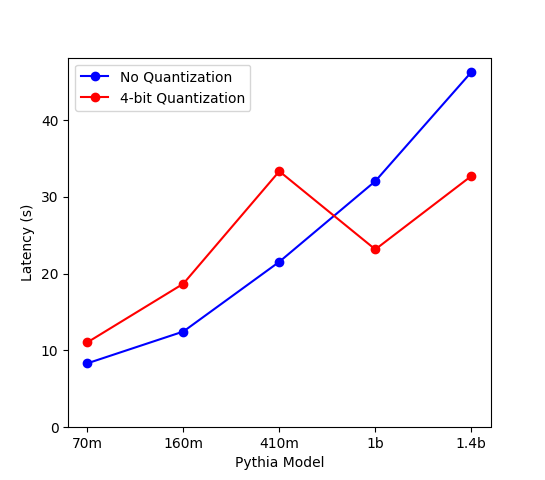}
    \vspace{-2mm}
    \caption{A comparison of the effects of quantization on median total token generation time for the Orin NX 16GB, at max NV power model.}
    \label{fig:quant_comp}
    \vspace{-6mm}
\end{figure}

\begin{figure}[h]
    \centering
    \vspace{-2mm}
    \includegraphics[trim={0 0 0 1.2cm},clip,width=1\linewidth, height=7cm]{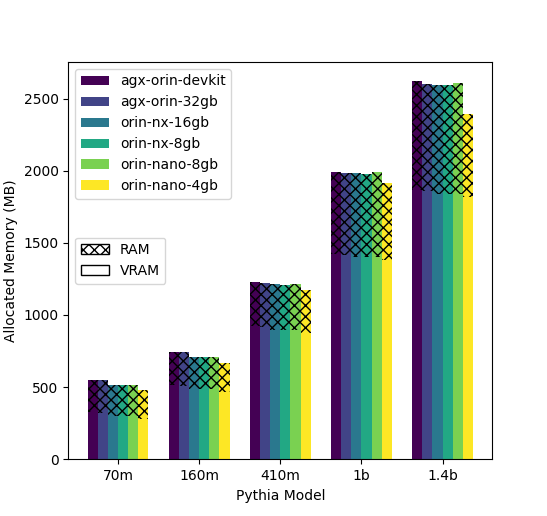}
    \vspace{-5mm}
    \caption{Median peak memory allocated during both model loading and token generation, at the max NV power model. Although the Jetson device configurations do not use separate memory hardware for RAM and VRAM, the distinction within the software is shown.}
    \label{fig:peak_mem}
    \vspace{-4mm}
\end{figure}

% JetPack 6.0, flash.sh
When preparing a new device configuration for testing, we used NVIDIA's `sdkmanager' application, which is provided to customers specifically for flashing Jetson devices\cite{sdkmanager}. This suite is usable in both a graphical and commandline format, and allows the user to flash a specific version of the JetPack SDK to their Jetson device. The JetPack SDK is an Ubuntu-based OS package with several libraries included that allow for more advanced AI packages to be run on the device.

For testing multiple configurations on the Jetson Orin developer kit, it was \textit{explicitly} required for us to run sdkmanager in an Ubuntu environment that would match the Ubuntu version to be installed on the device. For instance, when installing JetPack 6.0 (the latest JetPack version at the time of writing and the version we performed our tests on), we needed to prepare a host machine with Ubuntu 22.04. While NVIDIA does provide a Docker image to do this on other host operating systems, all our attempts at using this method failed (typically from errors in handling the physical device connection).

Each configuration could be flashed to the developer kit using a script inside sdkmanager called `flash.sh'. This script allows the user to pass different emulation configurations\cite{flash-sh} to the developer kit. This method generates a system image for the device based on the specifications of the desired configuration and flashes it to the device, provided that it is connected and booted into force recovery mode.

Because altering the device configuration of the Orin developer kit requires a complete re-flash of the device, it was not possible for our data and testing system to persist on the device between configurations.
Our solution to this was to use a single deployable GitHub repository, containing both the testing functionality as well as a modular setup script system. The setup system installs all necessary components required to run the tests (HuggingFace modules, PyTorch, etc.). Several of these require more specific installations than the typical method, such as needing to acquire a precompiled version of PyTorch from NVIDIA specifically for Jetson devices and compiling another of the HuggingFace modules (bitsandbytes) from source to compensate. On our device, the complete process for setup takes roughly 2 hours, so the setup process was made modular in case of any issues along the way. This allowed for the process to be restarted from roughly the point where it stopped instead of the beginning. Additionally, after data was collected, it was moved to a separate storage medium before the next flash.

Other methods exist for retaining the environment between configurations, such as adding an external or additional internal storage medium. The developer kit uses an internal eMMC storage system, but an M.2 storage medium may be installed as additional internal storage. External USB storage mediums are also usable with the developer kit. However, we opted against using any additional storage during our tests, on the grounds that adding hardware accessible at different speeds than the internal storage could affect our test results. Copying the entire environment between configurations was also ruled out because there may be issues with the environment when specific portions of the hardware are disabled/enabled between configurations.

\section{Results}

\subsection{Experimentation Results}

Out of a total of \textbf{210} possible configurations between our varied parameters (device configuration, quantization, NV power model, and LLM), our results cover \textbf{204} successful sets of iterations. Tables \ref{table:lat_load} and \ref{table:lat_gen} show the median latency across all the iterations for each of these successful configurations (failed configurations do not show any data). With five iterations each, the number of validated log files generated by our testing suite is \textbf{1,020}. Those that failed may have produced one or two valid logs, but were inevitably discarded due to the difficulty in reproducing the successes. The cause of these failures is discussed further in Section \ref{section:issues}.

The majority (unless otherwise specified) of our visualizations of our results use \textit{median} values across all iterations instead of the average. This is to avoid biasing by outliers due to the small number of iterations performed, as a few of the metrics show a first iteration with markedly different results from the rest of the iterations, which otherwise tend to fall very close between each other.

\vspace{0.5em}
\subsubsection{Shared Memory Issues}
\label{section:issues}

% memory is shared between cpu/gpu
As mentioned, Orin devices share a single bank of memory between the main system and the GPU \cite{unified_memory}. While this is beneficial in reducing the physical footprint of the SoC, our testing revealed limitations from attempting to use larger models with a fairly standard implementation. When using PyTorch as a backend for the HuggingFace libraries, memory is allocated as-needed. On a system with dedicated GPU hardware, this memory is allocated from VRAM; however, on these devices there is no physical distinction between RAM and VRAM. As such, PyTorch allocates directly from the same memory pool as the operating system.\\

While both PyTorch and HuggingFace provide methods for offloading portions of models to different memory locations (such as between VRAM and RAM, as well as the physical disk under high constraints), this functionality depends on multiple factors. An Out-Of-Memory (OOM) error for the VRAM of a dedicated GPU might not necessarily affect the general operation of the system; however, because of the unified memory of the Jetson devices, an OOM error by either portion of the SoC would have the same affect on the entire system.

% discuss failed nano 4gb tests and why
One particular issue during our testing involved running tests on the device configuration with the smallest amount of memory, the Orin Nano 4GB. When attempting to run the 1b and 1.4b Pythia models without quantization, the system would frequently reach an OOM error before the loading process completes. In these instances, the developer kit would complete freeze and remain unrecoverable until physically rebooted. In Tables \ref{table:lat_load} and \ref{table:lat_gen}, null entries are added where failed tests occured. In Figures \ref{fig:3d_gen_nq} and \ref{fig:3d_load_nq}, these are similarly shown as missing mesh points in the back corner.

Tests to determine the state of the system in these cases were inconclusive, but it neither HuggingFace/PyTorch or the Ubuntu system underneath were releasing memory, hinting that there may be some race condition preventing any memory allocations or deallocations.
Altering the Ubuntu operating system to reduce its memory footprint was considered out of scope for our testing; while it may be possible to create enough ``room'' for the 1b and 1.4b Pythia models to be loaded in, this may deviate from the standard usage of JetPack and the Jetson devices; it may alter the environment enough to produce data that is not indicative of these defaults.

The effects of varying swap was not tested, but the default configuration that comes with JetPack 6.0 has swap already enabled (and is reported by jtop). Despite this, we still experienced the issues above.

\subsubsection{Effects of Quantization}

% discuss interesting issue with quantization not affecting 1b and 1.4b the same way
All five Pythia models used in this study have been uploaded to HuggingFace with a 16-bit parameter precision, and we vary this between no quantization and 4-bit quantization (performed on-device). If we consider this 16-bit precision as a `baseline', we can use to to compare the effects of quantization on LLM performance on these devices. Figure \ref{fig:quant_comp} demonstrates this connection. The general consensus is that quantization allows for lower latency generation (at the cost of accuracy); however, out tests showed a significant \textit{increase} in latency for smaller models (70m - 410m) when quantized. The expected behavior returns for the larger models (1b, 1.4b) and quantization reduces the latency from the baseline. Figure \ref{fig:quant_comp} represents the Orin NX 16GB device configuration, but the same pattern can be seen in Figure \ref{fig:3d_gen_q} when sliced, showing a correlation across each device.

\begin{figure}[!t]
    \centering
    \vspace{4mm}
    \includegraphics[width=0.99\linewidth, height=7cm]{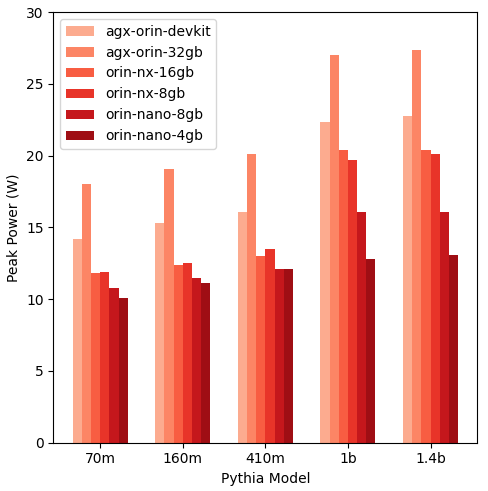}
    \caption{Median peak power usage (in watts) during token generation.}
    \label{fig:power}
    \vspace{-6mm}
\end{figure}

\begin{figure}[!t]
    \centering
    \includegraphics[width=0.99\linewidth, height=7cm]{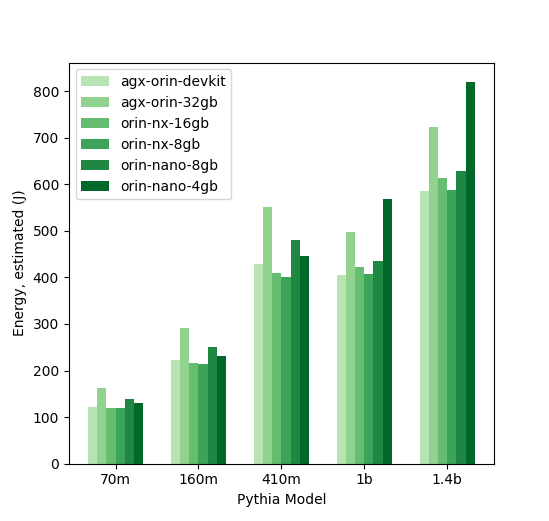}
    \caption{Estimated median energy usage (in joules) during total token generation, at the maximum NV power model and with 4-bit quantization.}
    \label{fig:energy}
    \vspace{-6mm}
\end{figure}

\begin{figure}[h]
    \centering
    \includegraphics[width=0.99\linewidth]{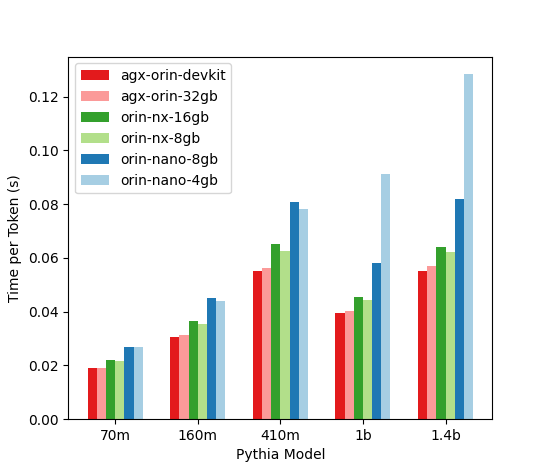}
    \caption{Average generation time per token, at the maximum NV power model of each configuration. The device ``families'' are highlighted by color.}
    \label{fig:tpt}
\end{figure}

\vspace{3mm}
\subsubsection{Other Results}

% general trend (70m has the least everything, 1.4b has the most)
In Figures \ref{fig:peak_mem}, \ref{fig:power}, and \ref{fig:energy}, the data shows a general trend for resource usage to increase as model size increases. This is understandable, as larger models require more allocated memory and perform more calculations than smaller models. However, there is \textit{not} a clear trend between device configurations across these metrics. In Figure \ref{fig:peak_mem}, the difference in peak allocated memory across device configurations is relatively level, whereas Figure \ref{fig:power} shows a considerable reduction in power as you move down to more constrained devices (with enough discrepancy that some readings are as low as half of others). 

% device families have similar results
One potential correlation we found in our test results is the effect of device configuration ``family'' on some metrics. Of the Orin configurations in this study, there are three device families: the AGX Orin, the Orin NX, and the Orin Nano. In Figure \ref{fig:tpt}, aside from the larger models on the Orin Nano 4GB, the time-per-token metric results are very close between members of the same family.

\subsection{``Use Cases'' for Constrained Applications}

For the purpose of application design, we can also use these results to determine the best configuration (Jetson Orin device, LLM, NV power model, and quantization) to use in your implementation, given a set of resource constraints. In each of the provided ``use cases'', examples of two separate constrained metrics have been chosen and are used to reduce the pool of test data, after which the highest/lowest of another specific metric is chosen. 

Additionally, an analysis script is provided that will auto-generate these for a given set of constraint data. The constraint data provided in these use cases is arbitrary.

\begin{table}[h]
    \vspace{0.4cm}
    \begin{subtable}[h]{0.45\textwidth}
        \centering
        \begin{tabular}{|c|c||c|}
            \hline
            \multicolumn{3}{|c|}{Use Case 1} \\
            \hline
            % Power & Accuracy & Configuration with Highest Accuracy \\ 
            % \hline
            % \multirow{2}{*}{$\leq$ 30 W} & \multirow{2}{*}{$\geq$ 50\%}
            % & AGX Orin Devkit, 30W NV power model, \\
            % & & pythia-1.4b-deduped, no quantization \\
            % \hline
            % \multirow{2}{*}{$\leq$ 20 W} & \multirow{2}{*}{$\geq$ 50\%}
            % & AGX Orin Devkit, 15W NV power model, \\
            % & & pythia-1.4b-deduped, no quantization \\
            % \hline
            % \multirow{2}{*}{$\leq$ 10 W} & \multirow{2}{*}{$\geq$ 40\%}
            % & Orin Nano 4GB, 7W-AI NV power model, \\
            % & & pythia-410m-deduped, 4-bit quantization \\
            % \hline
            Power & Latency & Configuration with Highest Accuracy \\ 
            \hline
            \multirow{2}{*}{$\leq$ 45 W} & \multirow{2}{*}{$\leq$ 40 s}
            & AGX Orin Devkit, 50W NV power model, \\
            & & pythia-1.4b-deduped, no quantization \\
            \hline
            \multirow{2}{*}{$\leq$ 30 W} & \multirow{2}{*}{$\leq$ 30 s}
            & AGX Orin Devkit, MAXN NV power model, \\
            & & pythia-1b-deduped, 4-bit quantization \\
            \hline
            \multirow{2}{*}{$\leq$ 15 W} & \multirow{2}{*}{$\leq$ 20 s}
            & Orin Nano 8GB, 15W NV power model, \\
            & & pythia-160m-deduped, no quantization \\
            \hline
        \end{tabular}
        \caption{Use Case \#1 Results}
        \label{table:uc1}
        \vspace{0.4cm}
    \end{subtable}
    \begin{subtable}[h]{0.45\textwidth}
        \centering
        \begin{tabular}{|c|c||c|}
            \hline
            \multicolumn{3}{|c|}{Use Case 2} \\
            \hline
            Energy & Pk. Memory & Configuration with Lowest Latency\\
            \hline
            \multirow{2}{*}{$\leq$ 240 J} & \multirow{2}{*}{$\leq$ 1400 MB}
            & AGX Orin Devkit, MAXN NV power model, \\
            & & pythia-70m-deduped, no quantization \\
            \hline
            \multirow{2}{*}{$\leq$ 240 J} & \multirow{2}{*}{$\leq$ 700 MB}
            & AGX Orin Devkit, MAXN NV power model, \\
            & & pythia-70m-deduped, 4-bit quantization \\
            \hline
            \multirow{2}{*}{$\leq$ 120 J} & \multirow{2}{*}{$\leq$ 700 MB}
            & Orin NX 8GB, 15W NV power model, \\
            & & pythia-70m-deduped, no quantization \\
            \hline
        \end{tabular}
        \caption{Use Case \#2 Results}
        \label{table:uc2}
        \vspace{0.4cm}
    \end{subtable}
    \begin{subtable}[h]{0.45\textwidth}
        \centering
        \begin{tabular}{|c|c||c|}
            \hline
            \multicolumn{3}{|c|}{Use Case 3} \\
            \hline
            Accuracy & Pk. Memory & Configuration with Lowest Latency \\
            \hline
            \multirow{2}{*}{$\geq$ 35\%} & \multirow{2}{*}{$\leq$ 800 MB}
            & AGX Orin Devkit, MAXN NV power model, \\
            & & pythia-70m-deduped, no quantization \\
            \hline
            \multirow{2}{*}{$\geq$ 45\%} & \multirow{2}{*}{$\leq$ 1200 MB}
            & AGX Orin 32GB, MAXN NV power model, \\
            & & pythia-160m-deduped, no quantization \\
            \hline
            \multirow{2}{*}{$\geq$ 55\%} & \multirow{2}{*}{$\leq$ 2000 MB}
            & AGX Orin Devkit, MAXN NV power model, \\
            & & pythia-1b-deduped, 4-bit quantization \\
            \hline
        \end{tabular}
        \caption{Use Case \#3 Results}
        \label{table:uc3}
    \end{subtable}
    \caption{Use case tables, given some example constraints.}
\end{table}

\vspace{0.5em}
\subsubsection{Use Case \#1}
Limited Power and Latency
% NOTE: changed from power and accuracy, as a minimum accuracy constraint when always returning the maximum accuracy seemed redundant

This first example use case takes a specified maximum power usage and latency to filter the data and produce the device configuration and LLM capable of the best accuracy.
See Table \ref{table:uc1} for specific information given our test results.

\vspace{0.5em}
\subsubsection{Use Case \#2}
Limited Energy and Memory

This example shows (based on our test results) what device configuration and LLM would have the lowest latency for a few specific energy usage and peak memory usage constraints.
See Table \ref{table:uc2} for specific information given our test results.

\vspace{0.5em}
\subsubsection{Use Case \#3}
Limited Accuracy and Memory

In this example, we use our results to determine what device configuration and LLM runs with the lowest latency, given some maximum peak memory and minimum accuracy constraints. 
See Table \ref{table:uc3} for specific information given our test results.

\section{Conclusions and Future Work}

The intention of this study is to provide benchmarking suite for LLM evaluation on embedded system for further research in hardware-software co-design and optimizations. Our testing suite is designed in such a way to promote customization, and is not limited to the Pythia LLMs that we targeted. One major limitation of our system (at the time of writing) is the reliance on jtop, which is only available for Jetson devices (though not limited to the Orin line). Performing LLM generation on non-Jetson devices was outside the scope of this work, but the logging system in our testing suite could be redesigned to use other statistics libraries.

%If accepted, the final revision of this paper will include a public GitHub link containing the testing suite developed during our study, as well as the recorded data.
The batch testing utility developed for our research is publicly accessible and can be found at: {\bf \url{https://github.com/LiamS57/orin-llm-testing}}. This repository includes all setup scripts necessary to prepare a flashed Orin device for testing, the testing utility itself, and a set of scripts for test data visualization. To duplicate our experiments with an Orin developer kit, the repository can be cloned onto the device after flashing it with the NVIDIA Jetson \textit{sdkmanager} utility.

One potential continuation for this work is to increase the iteration pool to obtain better estimates for our characterization. Although we perform multiple iterations of every parameter configuration, we only take 5 measurements per configuration in this study. Increasing the number of measurements would significantly increase the metric accuracy for a more robust baseline dataset.

\balance

%\section*{Acknowledgment}
% TODO: replace with the proper way to acknowledge aimslab and/or larri
%This work has been supported by the AIMSLab at the University of Louisville, and the US National Science Foundation (NSF REU \#2349076).
%This work has been supported by the \censor{AIMSLab} at the \censor{University of Louisville}, and the US National Science Foundation (NSF REU \censor{\#2349076}).

\bibliographystyle{IEEEtran}
\bibliography{ref}

\end{document}